\documentclass[letterpaper, 10 pt, conference]{ieeeconf}  

\IEEEoverridecommandlockouts                              

\usepackage{multirow, makecell}
\usepackage{hhline}
\usepackage{array, booktabs}
\usepackage{bm, amsmath}
\usepackage{balance}
\usepackage{graphicx} 
\usepackage{epsfig} 
\usepackage{float}
\usepackage[hidelinks]{hyperref}
\usepackage[linesnumbered,ruled,vlined]{algorithm2e}
\usepackage{algpseudocode}
\newcolumntype{P}[1]{>{\centering\arraybackslash}p{#1}}

\title{
	A Framework for Probabilistic Generic Traffic Scene Prediction 
}

\author{Yeping Hu, Wei Zhan and Masayoshi Tomizuka
\thanks{Y. Hu, W. Zhan and M. Tomizuka are with the Department of Mechanical Engineering, University of California, Berkeley, CA 94720 USA {\tt {[yeping\_hu, wzhan, tomizuka@berkeley.edu]}}}
}

\begin{document}

\maketitle
\thispagestyle{empty}
\pagestyle{empty}

\begin{abstract}
In a given scenario, simultaneously and accurately predicting every possible interaction of traffic participants is an important capability for autonomous vehicles. The majority of current researches focused on the prediction of an single entity without incorporating the environment information. Although some approaches aimed to predict multiple vehicles, they either predicted each vehicle independently with no considerations on possible interaction with surrounding entities or generated discretized joint motions which cannot be directly used in decision making and motion planning for autonomous vehicle. In this paper, we present a probabilistic framework that is able to jointly predict continuous motions for multiple interacting road participants under any driving scenarios and is capable of forecasting the duration of each interaction, which can enhance the prediction performance and efficiency. The proposed traffic scene prediction framework contains two hierarchical modules: the upper module and the lower module. The upper module forecasts the intention of the predicted vehicle, while the lower module predicts motions for interacting scene entities. An exemplar real-world scenario is used to implement and examine the proposed framework. 

\end{abstract}

\section{Introduction}
As the autonomous vehicle is becoming a big trend nowadays, safety is the most essential aspect to consider. Being able to predict future motions of the surrounding entities in the scene can greatly enhance the safety level of autonomous vehicles since potentially dangerous situations could be avoided in advance. Therefore, the Advanced Driver Assistance Systems (ADAS) is expected to have a full interpretation of the scene and be able to accurately predict possible behaviors of multiple traffic participants under various driving scenarios, which will then assure a safe, comfortable and cooperative driving experience. An illustration of the scene prediction is shown in Fig. 1.

There have been numerous works focusing on predicting the traffic participants in different driving scenarios. The predicted outcomes can be divided into two general categories : discrete and continuous. Intention estimation can be regarded as a discrete prediction problem, which is usually solved by classification strategies, such as Support Vector Machine (SVM) \cite{SVM_class}, Bayesian classifier \cite{BF_class}, Hidden Markov Models (HMMs) \cite{HMM_class}, and Multilayer Perceptron (MLP) 
\begin{figure}[htbp]
	\centering
	\includegraphics[scale=0.4]{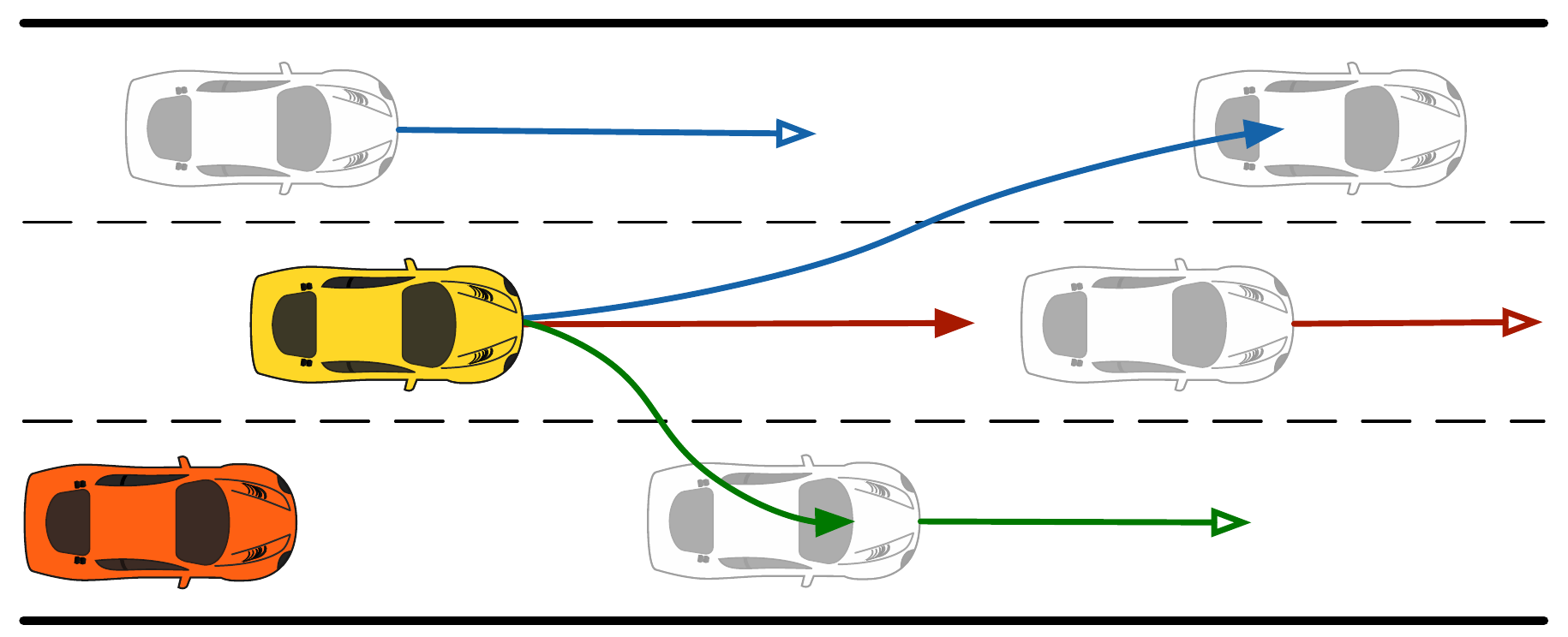}
	\caption{An illustration of the scene prediction results. The red car is the autonomous vehicle which is predicting the yellow car and its possible interaction with other scene entities.}
	\label{fig:scene_prediction}
\end{figure}
\cite{MLP_class}. Motion prediction, on the other hand, is mostly treated as a regression problem, where it forecasts the short-term movements and long-term trajectories of vehicles. Various motion prediction methods use vehicle kinematic models at the prediction step and estimate the state recursively. For example, methods such as constant velocity (CV), constant acceleration (CA) and Intelligent Driver Model (IDM) \cite{IDM} have been wildly used. However, these methods are generally considered in simple traffic scenarios such as car following. 

There are also many approaches that deal with motion predictions in more complicated situations such as lane changing and ramp merging. \cite{Traj_2} utilized Artificial Neural Network (ANN) structure to 
predict vehicle lateral motions and used a SVM to further determine if a lane change will happen. A Variational Gaussian Mixture Model (VGMM) is proposed in \cite{VGMM_2012} to classify and predict the long-term trajectories of vehicles in a simulated environment. The main limitation of these works, however, is that they failed to take surrounding vehicles into account, which is unreasonable since the trajectory of the predicted vehicle will be largely influenced by the environment.

To handle such problems, several works incorporated potentially affecting vehicles by making use of relational features. In \cite{LSTM_traj}, the authors used the long short-term memory (LSTM) to predict the most likely trajectories for vehicles in highway situation while considering their nine surrounding vehicles. \cite{MDN_2017} brought forward a Deep Neural Networks (DNN) to obtain the lateral acceleration and longitudinal velocity while taking into account five vehicles around the predicted car. Nevertheless, these methods only predict trajectories for a single scene entity without estimating the possible motions of other vehicles. Works such as \cite{Traj_4} and \cite{Traj_6} utilized a LSTM-based structure for each vehicle in the scene and predicted the probabilistic information on future locations over a occupancy grid map. Although possible trajectories for every surrounding vehicles were predicted, the authors treated each vehicle independently during the prediction process, which cannot provide sufficient and accurate predictions especially in highly interactive scenarios.

Very few studies have been done for simultaneously predicting multiple interacting traffic participants. The approaches of \cite{Interact_2} and \cite{Interact_4} do incorporate such interdependencies by jointly predicting behavior patterns of all on-road vehicles. However, acquiring only the discrete behavior pattern is not enough for autonomous vehicles to fully predict the traffic scene or to directly perform risk assessment.

All the aforementioned motion prediction works did not have an estimation for the prediction horizon. They either fixed the prediction length beforehand or recursively estimated the vehicle state until a designated location is reached. However, for a given scene, multiple types of interaction between vehicles are possible and each of them is expected to have different time span. Therefore, it is not only irrational but also computationally expensive to predict every vehicle trajectories for the same time horizon. 

In this paper, a probabilistic framework that is able to predict various types of dynamic scenes is proposed. It contains an upper module and a lower module, where the Semantic-based Intention and Motion Prediction (SIMP) method is used in the upper module and the Conditional Variational Autoencoder (CVAE) is used in the lower module. The upper module is capable of predicting possible semantic intention and motion of the selected vehicle, while the lower module can further predict joint probability distributions of motions for interacting traffic participants. Possible future trajectories will be sampled from each joint distribution and the prediction horizon for different interacting entities can be received from the upper module. Also, our framework can guarantee feasibilities for every sampled trajectory. These trajectories could then be easily used by the decision-making and motion planning process for autonomous vehicles.

The remainder of the paper is organized as follows: Section II provides the detailed explanation of the proposed scene prediction framework; Section III discusses an exemplar scenario to apply our framework; evaluations and results are provided in Section IV; and Section V concludes the paper. 

\section{Scene Prediction Framework}

In this section, we first introduce the method applied to each of the two modules in the scene prediction framework. Then, the overall prediction process of the framework is illustrated. 


\subsection{Upper Module}
For the upper module, we implemented the Semantic based Intention and Motion Prediction (SIMP) approach \cite{SIMP} due to its great adaptability for various scenarios and competitive prediction performance compared to other methods. It utilizes deep neural network to formulate a probabilistic framework which can predict the possible semantic intention and motion of the selected vehicle under various traffic scenes. The introduced semantics is defined as answering the question of \textit{"Which area will the predicted vehicle most likely insert into? Where and when?"}. The inserted area is called Dynamic Insertion Area (DIA), which can be a available gap between any two vehicles on the road or can be a lane entrance/exit area. Each DIA is assigned to a 2D Gaussian Mixture Model (GMM) to predict a two dimensional vector: $\bm{y} = [y_s, y_t]^T$, where $y_s$ describes vehicle's location and $y_t$ represents the time information. Note that both variables can be explicitly defined according to the problem formulation. 

Therefore, given a set of input features $\bm{x}$, the probability distribution $\bm{y}_a$ over a single area $a$ for the predicted vehicle can be expressed as: 
\begin{eqnarray}
p(\bm{y}_a|\bm{x}) =  \sum_{m=1}^{M}\alpha_{m}\frac{1}{2\pi{\sqrt{|\Sigma_m|}}}exp\left(-\frac{D_m^T\Sigma_m^{-1}D_m}{2}\right),
\end{eqnarray}
where $D_m = \bm{y}_a-\mu_m$ and $M$ denotes the total number of mixture components. For each mixture component $m$, the mixing coefficient $\alpha_{m}$, mean $\mu_m$, and covariance $\Sigma_m$ formulate a probability density function of the output $\bm{y}_a$.

The output of the SIMP structure contains the required parameters for every 2D GMM and the weight $w_a$ for each insertion area $a$. As for the desired outputs, not only the largest weight is expected to be associated to the actual inserted area, but also the highest probability is supposed to be at the proper location and time for the output distributions of that area. The loss function is then defined as
\begin{equation} \label{eq:loss}
\begin{split}
Loss &= W_1\bigg(-\sum_{n}log\bigg\{\sum_{a=1}^{N_a}\hat{w}_{a}^n p(\bm{y}_a^n|\bm{x})\bigg\}\bigg) \\
& \quad +W_2\bigg(-\sum_{n}\sum_{a=1}^{N_a}\hat{w}_{a}^nlog(w_{a}^n)\bigg),
\end{split} 
\end{equation}
where $N_a$ denotes the total number of DIA in the scene and $\hat{w}_{a}$ denotes the ground truth of the area weight, which is the one-hot-encoding of the final insertion area for the predicted vehicle. Parameters W1 and W2 are manually tuned such that the two loss components will have the same order of magnitude during training.

\subsection{Lower Module}

\begin{figure}[htbp]
	\centering
	\includegraphics[scale=0.26]{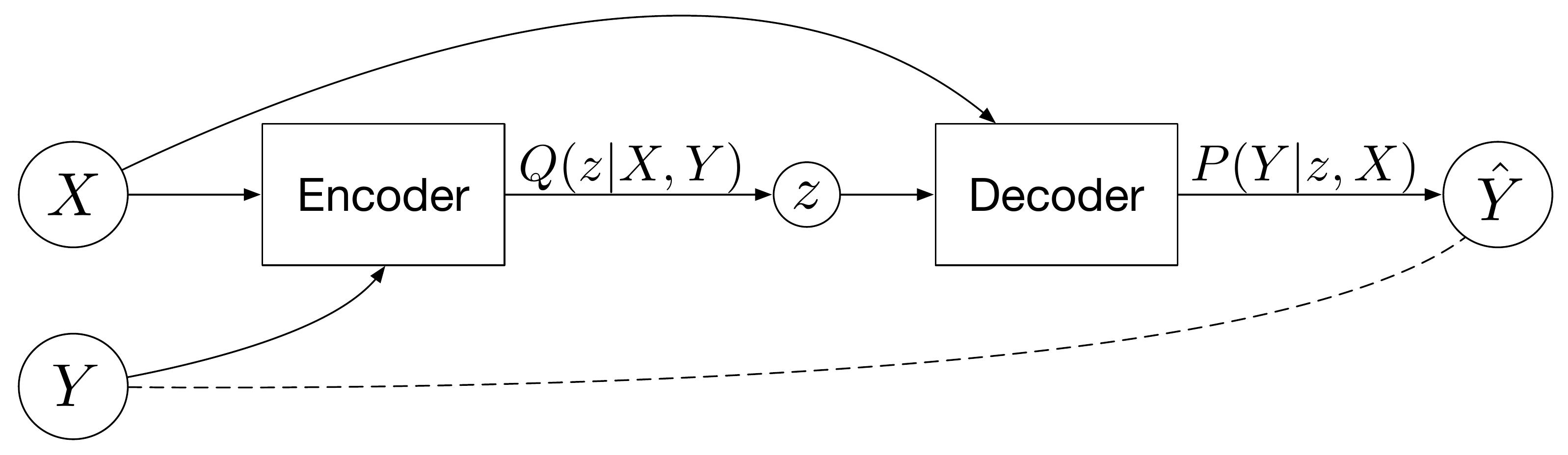}
	\caption{CVAE structure}
	\label{fig:CVAE}
\end{figure}

The lower module contains different motion models which will be assigned to each insertion area according to some criteria that will be discussed in Section III.
Here, we applied the Conditional Variational Autoencoder (CVAE) \cite{CVAE_1}\cite{CVAE_2} method to each motion model. 
CVAE has a similar structure to the typical variational autoencoder which contains an encoder and a decoder. It is rooted in bayesian inference, where the goal is to model the underlying probability distribution of the data so that new data could be sampled from that distribution. The overall structure of the CVAE we used is shown in Fig.~\ref{fig:CVAE}.


In order to obtain the distribution of the output $Y$ given the input $X$ (i.e. $P(Y|X)$), a latent variable $z \sim\mathcal{N}(0,I)$ is introduced such that 

\begin{eqnarray}
P(Y|X) = \mathcal{N}(f(z,X), \sigma^{2} \cdot I),
\end{eqnarray}
where it has mean $f$ which is a function that can be directly learned from the data and covariance equal to the identity matrix $I$ times a scalar $\sigma$.

Given the training data $(X,Y)$, the framework first samples $z$ from some arbitrary distribution $Q$ different from $\mathcal{N}(0,I)$. According to Bayes rule, we have: 
\begin{eqnarray}
\begin{split}
E_{z\sim Q}\big[\log{P(Y|z,X)}\big] &= E_{z\sim Q}\big[\log{P(z|Y,X)}\\ &+\log{P(Y|X)}-\log{P(z|X)}\big]
\end{split}
\end{eqnarray}
and by subtracting $E_{z\sim Q}(\log{Q(z)})$ from both sides:
\begin{eqnarray}
\begin{split}
\log{P(Y|X)}-E_{z\sim Q}\big[\log{Q(z)}-\log{P(z|X,Y)}\big] = \\ 
E_{z\sim Q}\big[\log{P(Y|z,X)}+\log{P(z|X)}-\log{Q(z)}\big].
\end{split}
\end{eqnarray}
To make the right hand side closely approximates $\log{P(Y|X)}$, $Q$ is constructed to depend on both $X$ and $Y$, which will make $E_{z\sim Q}\big[\log{Q(z)}-\log{P(z|X,Y)}\big]$ small. 

By writing the above expectation as Kullback-Leibler ($KL$) divergences, our variational objective is in the form:
\begin{eqnarray}
\begin{split}
\log{P(Y|X)}-D_{KL}\big[Q(z|X,Y)||P(z|X,Y)\big] = \\
E_{z\sim Q}\big[\log{P(Y|z,X)} - D_{KL}\big[Q(z|X,Y)||P(z|X)\big]\big],
\end{split}
\end{eqnarray}
where $P(z|X) = \mathcal{N}(0,I)$ since the model assumes that $z$ is independent of X if Y is unknown. The right hand side can be optimized via stochastic gradient descent by using the reparameterization trick to enable the encoder to generate a vector of means and a vector of standard deviation.  In general, Q is trained to "encode" Y into the latent $z$ space such that the values of $z$ can be "decoded" back to the output.

The loss function then becomes a summation of the generative loss, which is the mean square error between the network output and the ground truth, and a latent loss, which is the KL divergence term that forces the latent variables match a unit Gaussian:
\begin{eqnarray}
\begin{split}
Loss = ||Y - \hat{Y}||^2 + D_{KL}\big[Q(z|X,Y)||\mathcal{N}(0,I)\big],
\end{split}
\end{eqnarray}
where $Y$ is the ground truth and $\hat{Y}$ is the output estimation. At test time, we can sample from the distribution $P(Y|X)$ by sampling $z$ directly from $\mathcal{N}(0,I)$.
\begin{algorithm}
	\caption{Scene Prediction Framework}
	\label{alg:hybrid system}
	\SetAlgoLined
	\DontPrintSemicolon
	$N_s$ : total number of output samples\;
	$f_s$ : data sampling rate\;
	$\bm{x}$\quad : input vectors at current time \;
	$\bm{w}, p(y_s|\bm{x}), p(y_t|\bm{x}) \leftarrow  SIMP(\bm{x})$\;
	\For{$a = 1:N_a$}{
		$M$ $\leftarrow$ current motion model\;
		$n_a$ $\leftarrow$  $N_s \cdot w_a$\;
		\For{$i=1:n_a$}{
			$T$ $\leftarrow$ $sample( p(y_{t,a}|\bm{x}))$\;
			$\bm{s}_1,\bm{o}_1$ $\leftarrow$ $\bm{x}$\;
			\For{$j=1:(T \cdot f_s)$}{
				$T_{j}$ $\leftarrow$ $T - (j-1) \cdot f_s$\;
				$\bm{a}_j$ $\leftarrow$ $sample(M(\bm{s}_j,\bm{o}_j,T_{j}))$\;
				\eIf{$\bm{a}_j$ not feasible}{
					redo iteration\;}
				{$\bm{s}_{j+1}$ $\leftarrow$ $f(\bm{a}_j,\bm{s}_j)$\;}
			}
			\If{$p(y_{s,a} = \bm{s}_{final}|\bm{x}) < \epsilon$}{
				redo iteration\;
			}
		}
		
	}	
\end{algorithm}

\subsection{The Overall Framework}

For a selected vehicle, its input feature vector $\bm{x}$ contains all the information of itself and its surrounding environment. The SIMP method takes $\bm{x}$ as the input and generates three types of outputs: $\bm{w}$, $p(y_s|\bm{x})$, and $p(y_t|\bm{x})$. The vector $\bm{w}$ represents the weight for each of the $N_a$ areas; $p(y_s|\bm{x})$ denotes the probabilistic distribution of the final destination within each DIA; and $p(y_t|\bm{x})$ denotes the probabilistic distribution of the time remained to enter each DIA for the predicted vehicle, which can be also interpreted as the time-to-lane-change (TTLC) distribution. These outputs along with the filtered input features will then feed into different motion models inside the lower module. Finally, the joint probability distribution of motions over multiple interacting entities at the next time step can be obtained. At the current time step $t$, we denote states of the interacting vehicles as $\bm{s}_t$, the other sensor observations of the surroundings as  $\bm{o}_t$, and the predicted actions as $\bm{a}_t$. The next state of the interacting vehicles can be directly calculated using some mapping function $f$ such that $\bm{s}_{t+1} = f(\bm{a}_t, \bm{s}_{t})$.

Therefore, given the current state information and according to the first-order Markov assumption, the joint probability distribution over the prediction horizon $T$ can be expressed as: 
\begin{eqnarray}
\begin{split}
p(\bm{s}_{t+1},\bm{s}_{t+2},...,\bm{s}_{t+T}|\bm{s}_t) &=   p(\bm{s}_{t+1}|\bm{s}_{t})p(\bm{s}_{t+2}|\bm{s}_{t+1})\\
& \quad ...p(\bm{s}_{t+T}|\bm{s}_{t+T-1}),
\end{split}
\end{eqnarray}
where $T$ has the distribution of $p(y_t|\bm{x})$. Since $\bm{s}_t$ is independent of $\bm{o}_t$ and $T$, we have:
\begin{eqnarray}
\begin{split}
p(\bm{s}_{t+1}|\bm{s}_{t})& = \sum_{T}\sum_{\bm{o}_t}p(\bm{s}_{t+1},\bm{o}_t,T|\bm{s}_t)\\
& = \sum_{T}\sum_{\bm{o}_t}p(\bm{s}_{t+1}|\bm{s}_t,\bm{o}_t,T)p(\bm{s}_t),
\end{split}
\end{eqnarray}
where $p(\bm{s}_{t+1}|\bm{s}_t,\bm{o}_t,T)$ is obtained from the output distribution of the motion prediction model and $\bm{o_t}$ is assumed to have a gaussian measurement noise. Here, we use sample-based method to infer the desired joint probability distribution and the overall process of the proposed scene prediction framework is illustrated in Algorithms~\ref{alg:hybrid system}.
\begin{figure*}[htbp]
	\centering
	\includegraphics[scale=0.32]{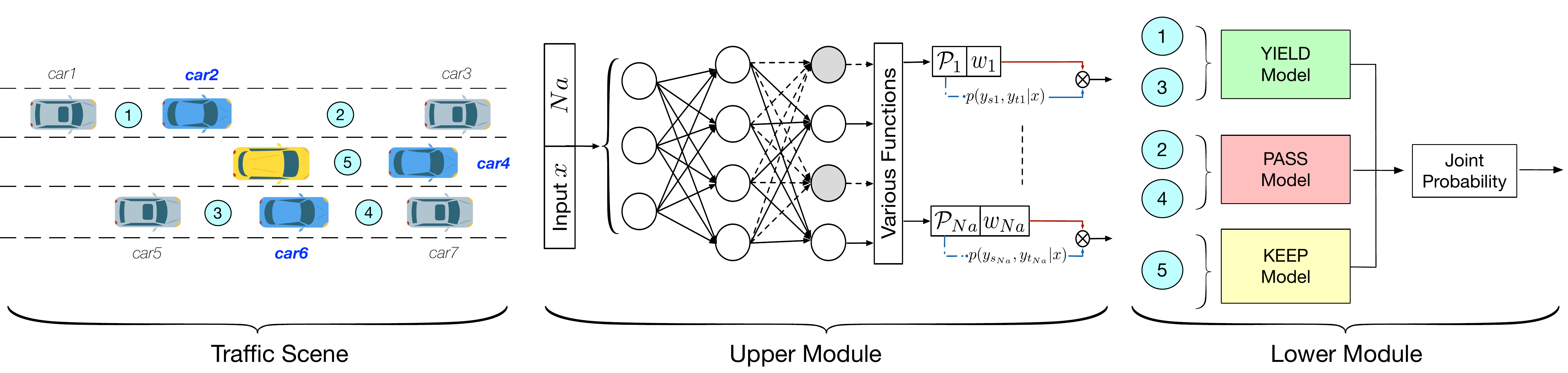}
	\caption{The scene prediction framework for the selected exemplar scenario.}
	\label{fig:hybrid_system}
\end{figure*}

Each motion model has its corresponding dynamic insertion area and the model will be triggered once its insertion weight $w_a$ is greater than a threshold (i.e. $n_a$ is not zero). Within each trajectory sampling process, we first sample the time information $T$, from $p(y_t|\bm{x})$. Then, within the given prediction horizon $T \cdot f_s$, we iteratively apply our lower motion model to obtain the sampled output action, which will go through a feasibility check to make sure the predicted next state is reachable by the vehicle. Finally, we use the destination distribution $p(y_s|\bm{x})$ from the SIMP method to ensure our predicted final state is within the desired range. 


%

\section{An Exemplar Scenario}
In this section, we use an exemplar  highway driving scenario to apply our proposed scene prediction framework. The data source and detailed problem formulation are presented. 
\subsection{Dataset}

The data we used was taken from the NGSIM dataset which is publicly available online at \cite{NGSIM}. We used the US highway 101 dataset which contains detailed vehicle trajectory data collected with sampling rate $f_s = 10$ Hz. Since the original data, especially for the velocity and acceleration, are very noisy, we used an Extended Kalman Filter (EKF) for better estimations. For lane-change situation, we picked up to 40 frames (4s) before the vehicle's center intersects the lane mark; for lane-keep situation, 40 frames were selected for each vehicle. A total of 15,240 frames were chosen from the dataset and randomly split into 80\% for training and 20\% for testing.

\subsection{Scenarios and Problem Description}
The representation of the exemplar scenario and the corresponding framework structure are shown in Fig.~\ref{fig:hybrid_system}. The yellow car is the predicted vehicle; the blue cars are the reference vehicles, which are vehicles that the predicted vehicle will most likely interact with; the grey cars are other surrounding vehicles that are assumed to be fully observable.

In the typical driving scenario, there are five circled areas (DIA) that our predicted vehicle could eventually enter. More specifically, if the predicted vehicle (yellow car) inserts into $area1$ or $area3$, it will yield to car2 or car6 respectively; if the predicted vehicle enters $area2$ or $area4$, it will pass car2 or car6 respectively; if, however, $area5$ is inserted, the predicted vehicle will keep its lane and follow car4. Therefore, we considered three motion models inside the lower module: \textbf{yield}, \textbf{pass} and \textbf{keep}. Each model is trained separately on data that satisfy each particular model type. Note that for other traffic scenes such as ramp merging, the same type of motion models can be directly applied.

\subsection{Features and Structure Details }
The input features for the upper module are selected as same as in \cite{SIMP}. For the lower module, input states of the two interactive vehicles can be written as $\bm{s}_t = \{v_{pred}^t,v_{ref}^t,x_{ref}^t-x_{pred}^t,y_{ref}^t-y_{pred}^t\}$, where the subscript $pred$ denotes the predicted vehicle and $ref$ denotes the reference vehicle, and the observed input $\bm{o}_t$ is determined by the type of the motion model. For example, $\bm{o}_t$ for the \textit{pass} model contains states of the vehicle in front of the predicted and the reference car, whereas no $\bm{o}_t$ is needed for the \textit{keep} model since only the front reference car need to be considered which information is already included in the input state. The predicted actions $\bm{a}_t$ for each motion model are expressed as $\bm{a}_t = \{\Delta{x}_{pred}^{t}, \Delta{v}_{pred}^{t},\Delta{x}_{ref}^{t}, \Delta{v}_{ref}^{t}\}$, representing lateral displacements and the longitudinal velocity differences for both the predicted vehicle and the reference vehicle. Here, the aforementioned function $f$ will map $\bm{a}_t$ and $\bm{s}_t$ linearly to $\bm{s}_{t+1}$.
\begin{figure*}[htbp]
	\centering
	\includegraphics[scale=0.4]{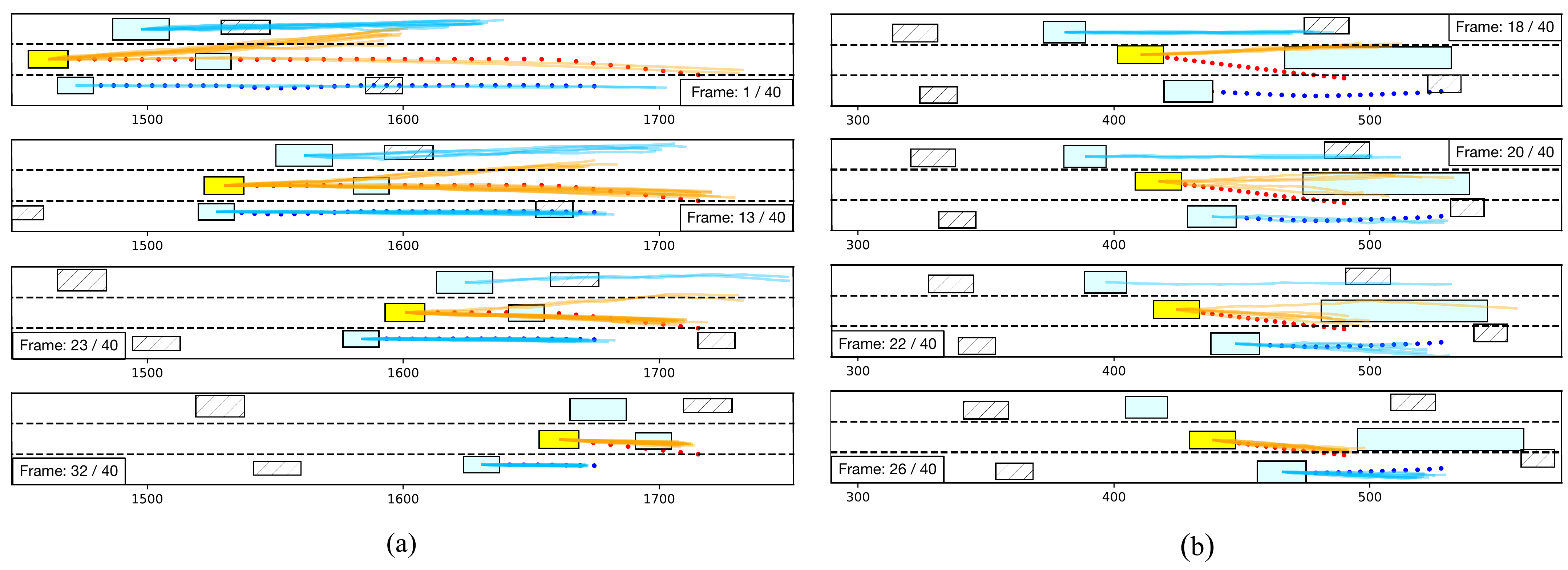}
	\caption{Test results of two example cases. The yellow car represents the predicted vehicle; the blue cars are reference vehicles that might interact with the predicted vehicle; and the striped cars denote other surrounding vehicles which will not have direct interactions with the predict vehicle. The red and blue dotted lines are the ground truth trajectories (with 0.1s step-time) for the predicted vehicle and its interacting reference vehicle respectively. The solid lines are sampled trajectories from the motion models.}
	\label{fig:visualization}
\end{figure*}

For the SIMP structure, we use three fully connected layers of 400 neurons each, with $tanh$ non-linear activation functions. A dropout layer is appended to the end to enhance the network's generalization ability and prevent overfitting. For the CAVE structure, we utilize two fully connect layers of 128 neurons each for both the encoder and the decoder, and use three latent variables. The two modules are trained hierarchically and concatenated during testing. 

\section{Evaluation and Results}
In this section, different evaluation methods are presented to assess the model quality and the final results are discussed.

\subsection{Performance Evaluation}
The output of the scene prediction framework are trajectories sampled from joint motion distributions of the predicted vehicle and each of its interacting entities (i.e. reference vehicles). We used the Root-Mean-Square Error (RMSE) as the validation metric to evaluate the output trajectories. However, multiple possible trajectories can be sampled from different motion models at the early stage of a lane changing scenario and more trajectories will be sampled from the ground-truth motion model as the time-to-lane-change value gets smaller. Therefore, in order to calculate the RMSE for a longer horizon, we analyzed our result by sampling trajectories directly from the actual motion model during the whole prediction period and compared them with the real trajectory. 

For a selected test case, the ground-truth motion model is determined by the final insertion area and the RMSE can be calculates as:
\begin{equation}
RMSE = \sqrt{\dfrac{1}{N_s \cdot T \cdot f_s}\sum_{i=1}^{N_s}\sum_{j=1}^{T \cdot f_s}(\bm{s}_j-\hat{\bm{s}_j^i})},
\end{equation} 

where $\bm{s}_j$ is the true state value at time $j$ and $\hat{\bm{s}_j^i}$ is the predicted value from the $i$-th sampled trajectory at the same time step $j$. We evaluated the RMSE of the lateral position and the longitudinal velocity separately for each motion model. Moreover, we selected the \textit{pass} model to evaluate the performance of using and not using the time-to-lane-change information obtained from the upper module. The RMSE results across different motion models under various prediction horizons are presented in Table~\ref{tab:RMSE}

As shown in Table~\ref{tab:RMSE}, our method has lateral position errors within 0.5m for each model and a 1.3m/s error in velocity when prediction horizon is 4s. For the lateral position error, it can be clearly seen that the performance of the reference vehicle is better than that of the predicted vehicle. This is because during the lane changing period, the predicted car has larger lateral displacement than the reference car which usually remains its lateral position. When the \textit{pass} model does not use the time information given by the SIMP method, the RMSE values increase especially for velocity errors. This happened because without the time constraint, the model may assume both vehicles will remain their speed and none of them will yield to the other at the beginning stage, which could cause cumulated speed deviations for both vehicles as well as a delay in the lateral displacement for the predict vehicle (yellow). Note that only a 4s prediction horizon is examined for the \textit{keep} model since there won't have much motion changes within that period for car following cases.

\begin{table}[ht]
	\centering
	\caption{RMSE Evaluation Results}
	\label{tab:RMSE}
	\begin{tabular}{P{1.2cm} P{0.8cm} P{0.5cm} P{0.5cm} P{0.5cm} P{0.5cm} P{0.5cm} P{0.5cm}}
		\firsthline
		\noalign{\smallskip}
		&&\multicolumn{6}{c}{Prediction horizon} \\
		\cline{3-8} 
		\noalign{\smallskip}
		Model & Vehicle & 4s & 3s & 2s & 1.5s & 1s & 0.5s \\
		\midrule
		\multirow{2}{*}[-0.1cm]{\textbf{Keep}} & $pred$ & 0.30 & - & - & - & - & - \\[1.2mm]
		& $ref$ & 0.28 & - & - & - & - & - \\
		\midrule
		\multirow{2}{*}[-0.1cm]{\textbf{Yield}} & $pred$ & 0.51 & 0.48 & 0.39 & 0.21 & 0.14 & 0.08 \\[1.2mm]
		& $ref$ & 0.42 & 0.36 & 0.25 & 0.18 & 0.11 & 0.07 \\
		\midrule
		\multirow{2}{*}[-0.1cm]{\textbf{Pass}} & $pred$ & 0.43 & 0.41 & 0.37 & 0.21 & 0.19 & 0.07 \\[1.2mm]
		& $ref$ & 0.41 & 0.33 & 0.22 & 0.14 & 0.04 & 0.02 \\
		\midrule
		\multirow{2}{*}[-0.1cm]{\shortstack[lb]{\textbf{Pass}\\ (w/o time)}} & $pred$ & 1.10 & 0.77 & 0.67 & 0.51 & 0.28 & 0.12 \\[1.2mm]
		& $ref$ & 0.34 & 0.30 & 0.29 & 0.18 & 0.07 & 0.03 \\
		\midrule 
	\end{tabular}
	\\[1.2mm] (a) RMSE for Lateral Position (m) \\
	\bigskip
	
	\begin{tabular}{P{1.2cm} P{0.8cm} P{0.5cm} P{0.5cm} P{0.5cm} P{0.5cm} P{0.5cm} P{0.5cm}}
		\firsthline
		\noalign{\smallskip}
		&&\multicolumn{6}{c}{Prediction horizon} \\
		\cline{3-8} 
		\noalign{\smallskip}
		Model & Vehicle & 4s & 3s & 2s & 1.5s & 1s & 0.5s \\
		\midrule
		\multirow{2}{*}[-0.1cm]{\textbf{Keep}} & $pred$ & 1.11 & - & - & - & - & - \\[1.2mm]
		& $ref$ & 1.07 & - & - & - & - & - \\
		\midrule
		\multirow{2}{*}[-0.1cm]{\textbf{Yield}} & $pred$ & 1.24 & 1.19 & 0.91 & 0.79 & 0.70 & 0.31 \\[1.2mm]
		& $ref$ & 1.25 & 1.05 & 0.72 & 0.67 & 0.33 & 0.22 \\
		\midrule
		\multirow{2}{*}[-0.1cm]{\textbf{Pass}} & $pred$ & 1.23 & 1.08 & 0.89 & 0.74 & 0.50 & 0.25 \\[1.2mm]
		& $ref$ & 1.34 & 1.01 & 0.74 & 0.58 & 0.49 & 0.20 \\
		\midrule
		\multirow{2}{*}[-0.1cm]{\shortstack[lb]{\textbf{Pass}\\ (w/o time)}} & $pred$ & 2.59 & 2.25 & 1.97 & 1.41 & 0.90 & 0.79 \\[1.2mm]
		& $ref$ & 2.17 & 2.06 & 1.32 & 1.08 & 0.64 & 0.43 \\
		\midrule
	\end{tabular}
	\\[1.2mm] (b) RMSE for Longitudinal Velocity (m/s)\\
\end{table}

\subsection{Visualization of Selected Cases}
We selected two distinct traffic situations to visualize our results as shown in Fig.~\ref{fig:visualization}. Each situation has a maximum prediction horizon of 4s (40 frames) and four representative frames are selected from each scenario to illustrate the performance of our framework. In order to better visualize the results, we did not add the trajectories generated by the \textit{keep} motion model on the plot. 

At the early stage of the situation in Fig.4 (a), the predicted vehicle is likely to interact with its left reference car since the vehicle is closer to the area behind its left car than the area in front of its right car. Therefore, in the first frame, most trajectories are sampled from the \textit{yield} motion model and the yielding pattern can be clearly recognized from the plot. As time goes, the predicted vehicle starts to increase its speed and finally passes its right reference vehicle. For case in Fig.4 (b), the predicted vehicle changes its mind by first having a large intention to pass its left vehicle and then choosing to yield to its right reference vehicle. Note that both cases have some intermediate stages where multiple possible interaction exist and thus it is necessary to predict all these potential joint trajectories to have a full understanding of the scene.

\subsection{Framework Robustness Test}
Although our framework shows great performances during the evaluation, it is possible that some corner cases are not included in the test data. Therefore, to examine whether our method can generate reasonable results under various situations, we manually changed some parameter settings to have an comprehensive analysis. We used two methods to evaluate the framework robustness: 
\begin{itemize}
	\item Fix the current input scenario and change motion models in the lower module.
	\item Fix the motion model in the lower module and change the time-to-lane-change (TTLC) distribution obtained from the upper module.
\end{itemize}

\begin{figure}[htbp]
	\centering
	\includegraphics[scale=0.35]{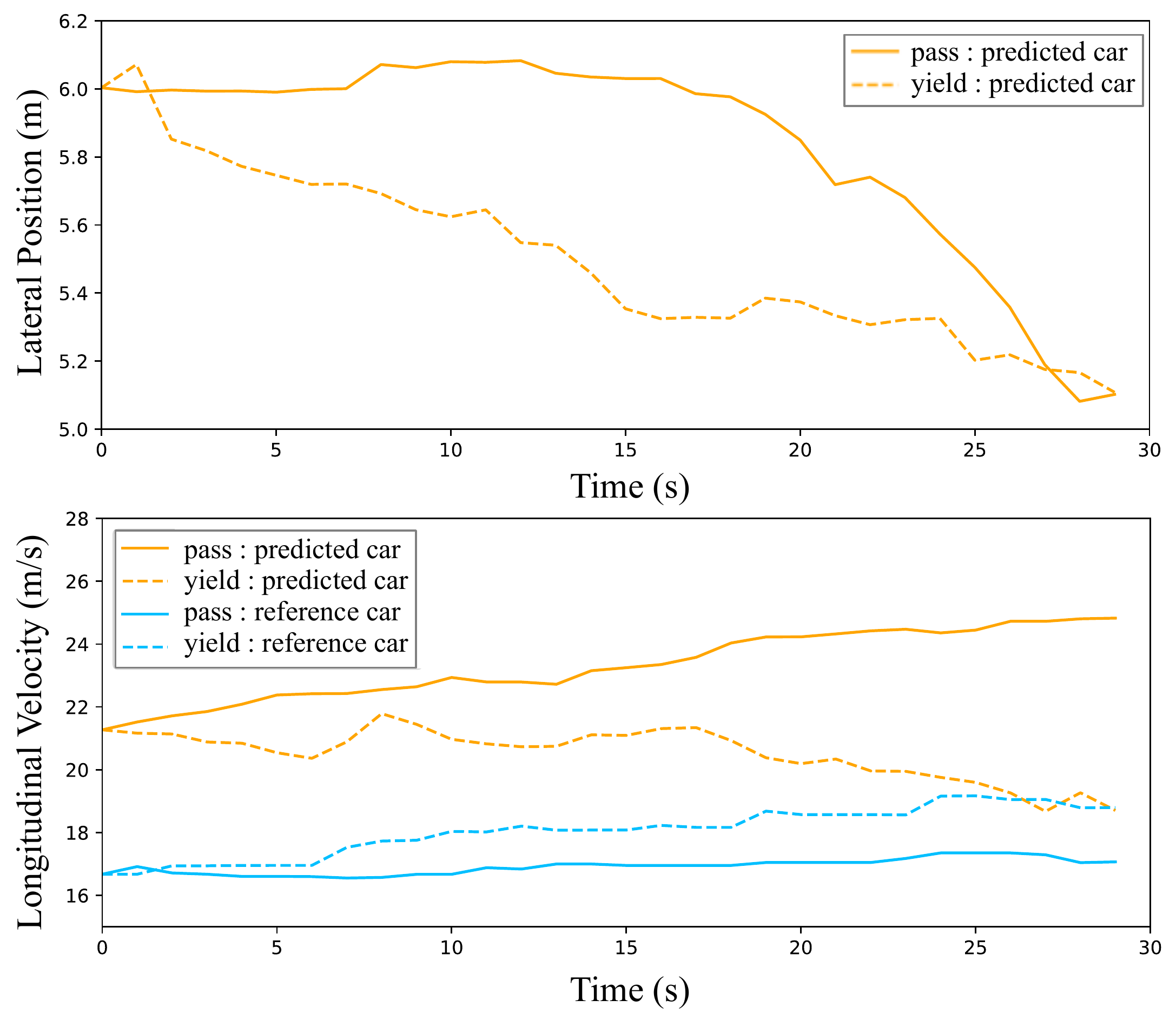}
	\caption{Performance Comparison using Different Motion Models}
	\label{fig:change_model}
\end{figure}

For the first evaluation, we chose the 10th frame of the test scenario in Fig.~\ref{fig:visualization} (a), where the predicted car has a similar longitudinal position as its right reference car and the actual motion model is \textit{pass}. We then changed the motion model to \textit{yield} and the comparison results of using different models are shown in Fig.~\ref{fig:change_model}. As can be seen in the plot, the \textit{pass} model predicts that the predicted vehicle will first increase the speed to surpass the reference vehicle and then perform large lateral deviations for lane change. The \textit{yield} model, however, indicates that the predicted vehicle will decrease its speed at the beginning and yield the reference car while changing its lane. Also, as the model is switched from \textit{pass} to \textit{yield}, the longitudinal velocity of the reference vehicle increases since the predicted car is assumed not to insert in front of it due to the decreasing speed of the predicted car.

\begin{figure}[htbp]
	\centering
	\includegraphics[scale=0.35]{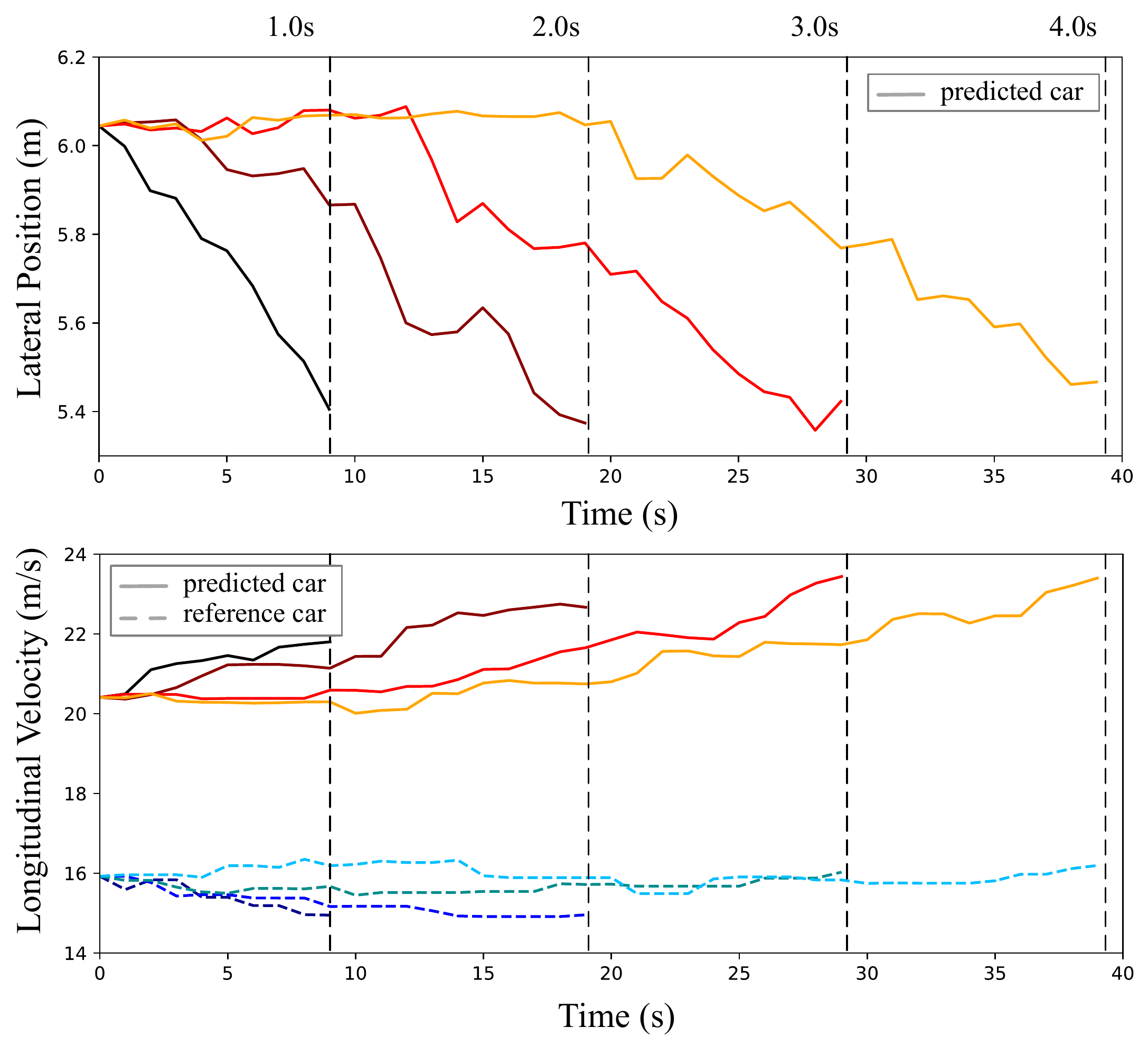}
	\caption{Performance Comparison using Different Temporal Information}
	\label{fig:change_ttlc}
\end{figure}

We used the same test scenario for the second evaluation but evaluated from the starting frame. By raising the predicted TTLC value from 1s to 4s, the predicted vehicle will have more time to change its lane. According to Fig.~\ref{fig:change_ttlc}, when TTLC is 1s, the predicted car has a 0.6m lateral deviation within one second; when TTLC increases, the vehicle gradually changes its lane, which takes up to 2s to have the same amount of lateral deviation. The speed comparison also reasonably indicates that the vehicle will have a larger acceleration when it decides to change the lane faster. 

\section{Conclusions}
In this paper, a framework for probabilistic traffic scene prediction was proposed. It can simultaneously predict possible motions for multiple interacting traffic participants under various circumstances. An exemplar highway scenario with real-world data was used to demonstrate the performance of the framework. The proposed framework is not only able to generate accurate trajectories sampled from the predicted joint distributions of scene entities, but also has the adaptability to different driving situations. Note that we used the CVAE method to analyze the performance of the framework and it does not necessarily conclude that this is the best suited to the lower module. Other approaches that can generate conditional joint distributions are also able to use and we will compare different methods as well as examine the framework on other scenarios in our future works. 

\balance

\end{document}